\documentclass{article}

\usepackage{microtype}
\usepackage{graphicx}
\usepackage{subfigure}
\usepackage{booktabs} 
\usepackage{enumitem}
\usepackage{multicol}
\usepackage{makecell}

\usepackage{hyperref}

\usepackage[table]{xcolor}

\newcommand\nohyph{\hyphenpenalty=10000\relax\exhyphenpenalty=10000\relax} 

\usepackage[accepted]{icml2019}

\icmltitlerunning{The Evolved Transformer}

\hypersetup{draft} 
\begin{document}

\twocolumn[
\icmltitle{The Evolved Transformer}

\icmlsetsymbol{equal}{*}

\begin{icmlauthorlist}
\icmlauthor{David R. So}{goob}
\icmlauthor{Chen Liang}{goob}
\icmlauthor{Quoc V. Le}{goob}
\end{icmlauthorlist}

\icmlaffiliation{goob}{Google Research, Brain Team, Mountain View, California, USA}

\icmlcorrespondingauthor{David R. So}{davidso@google.com}

\icmlkeywords{Evolution, Neural Architecture Search, Transformer, NLP}

\vskip 0.3in
]

\printAffiliationsAndNotice{}

\begin{abstract}
Recent works have highlighted the strength of the \textit{Transformer} architecture on sequence tasks while, at the same time, \textit{neural architecture search} (NAS) has begun to outperform human-designed models. Our goal is to apply NAS to search for a better alternative to the Transformer. We first construct a large search space inspired by the recent advances in feed-forward sequence models and then run evolutionary architecture search with \textit{warm starting} by seeding our initial population with the Transformer. To directly search on the computationally expensive WMT 2014 English-German translation task, we develop the \textit{Progressive Dynamic Hurdles} method, which allows us to dynamically allocate more resources to more promising candidate models. The architecture found in our experiments -- the \textit{Evolved Transformer} -- demonstrates consistent improvement over the Transformer on four well-established language tasks: WMT 2014 English-German, WMT 2014 English-French, WMT 2014 English-Czech and LM1B. At a big model size, the Evolved Transformer establishes a new state-of-the-art BLEU score of 29.8 on WMT'14 English-German; at smaller sizes, it achieves the same quality as the original "big" Transformer with 37.6\% less parameters and outperforms the Transformer by 0.7 BLEU at a mobile-friendly model size of $\sim$7M parameters.

\end{abstract}

\setlength{\textfloatsep}{20pt}
\setlength{\intextsep}{20pt}

\section{Introduction}

Over the past few years, impressive advances have been made in the field of neural architecture search. Reinforcement learning and evolution have both proven their capacity to produce models that exceed the performance of those designed by humans \cite{real19, zoph18}. These advances have mostly focused on improving vision models, although some effort has also been invested in searching for sequence models \cite{zoph16, pham18}. In these cases, it has always been to find improved recurrent neural networks (RNNs), which were long established as the de facto neural model for sequence problems \cite{sutskever14,bahdanau2015}.

However, recent works have shown that there are better alternatives to RNNs for solving sequence problems. Due to the success of convolution-based networks, such as Convolution Seq2Seq~\cite{gehring2017}, and full attention networks, such as the Transformer \cite{vaswani17}, feed-forward networks are now a viable option for solving sequence-to-sequence (seq2seq) tasks. The main strength of feed-forward networks is that they are faster, and  easier to train than RNNs.

The goal of this work is to examine the use of neural architecture search methods to design better feed-forward architectures for seq2seq tasks.   
Specifically, we apply \textit{tournament selection} architecture search and \textit{warm start} it with the Transformer, considered to be the state-of-art and widely-used, to evolve a better and more efficient architecture. To achieve this, we construct a search space that reflects the recent advances in feed-forward seq2seq models and develop a method called \textit{Progressive Dynamic Hurdles} (PDH) that allows us to perform our search directly on the computationally demanding WMT 2014 English-German (En-De) translation task. Our search produces a new architecture -- called the \textit{Evolved Transformer} (ET) -- 
which demonstrates consistent improvement over the original Transformer on four well-established language tasks: WMT 2014 English-German, WMT 2014 English-French (En-Fr), WMT 2014 English-Czech (En-Cs) and the 1 Billion Word Language Model Benchmark (LM1B). At a big model size, the Evolved Transformer establishes a new state-of-the-art BLEU score of 29.8 on WMT'14 En-De. It is also effective at smaller sizes, achieving the same quality as the original "big" Transformer with 37.6\% less parameters and outperforming the Transformer by 0.7 BLEU at a mobile-friendly model size of $\sim$7M parameters.

\section{Related Work}

RNNs have long been used as the default option for applying neural networks to sequence modeling \cite{sutskever14,bahdanau2015}, with LSTM \cite{hochreiter97} and GRU \cite{cho14} architectures being the most popular. However, recent work has shown that RNNs are not necessary to build state-of-the-art sequence models. For example, many high performance convolutional models have been designed, such as WaveNet~\cite{van2016wavenet}, Gated Convolution Networks~\cite{dauphin2017language}, Conv Seq2Seq~\cite{gehring2017} and the Dynamic Lightweight Convolution model~\cite{wu2018pay}. Perhaps the most promising architecture in this direction is the Transformer architecture~\cite{vaswani17}, which relies only on multi-head attention to convey spatial information. In this work, we use both convolutions and attention in our search space to leverage the strengths of both of these layer types. 

The recent advances in sequential feed-forward networks are not limited to architecture design. Various methods, such as BERT \cite{devlin18} and Radford et. al's \yrcite{radford18} pre-training technique, have demonstrated how models such as the Transformer can improve over RNN pre-training~\cite{dai2015semi,Peters:2018}.
For translation specifically, work on scaling up batch size \cite{ott18, wu2018pay}, using relative position representations \cite{shaw18}, and weighting multi-head attention \cite{ahmed17} have all pushed the state-of-the-art for WMT'14 En-De and En-Fr. However, these methods are orthogonal to this work, as we are only concerned with improving the neural network architecture itself, and not the techniques used for improving overall performance.

The field of neural architecture search has also seen significant recent progress. The best performing architecture search methods are those that are computationally intensive \cite{zoph16, baker2016designing, real17, Xie2017GeneticC, zoph18, real19}. Other methods have been developed with speed in mind, such as DARTS \cite{hliu18}, ENAS \cite{pham18}, SMASH \cite{brock2018smash}, and SNAS \cite{xie19}. These methods radically reduce the amount of time needed to run each search by approximating the performance of each candidate model, instead of investing resources to fully train and evaluate each candidate separately. However, these methods also have several disadvantages that make them hard to apply in our case: (1) It is hard to warm start these methods with the Transformer, which we found to be necessary to yield strong results. (2) ENAS and DARTS require too much memory at the model sizes we are searching for. (3) The best architecture in the vision domain (e.g., AmoebaNet\cite{real19}) was discovered by evolutionary NAS, not these efficient methods, and we optimize for best architecture over best search efficiency here.

Zela et. al's \yrcite{zela18} utilization of Hyperband \cite{li17} and PNAS's \cite{liu18} incorporation of a surrogate model are examples of approaches that try to both increase efficiency via candidate performance estimation and maximize search quality by training models to the end when necessary. The Progressive Dynamic Hurdles (PDH) method we introduce here is similar to these approaches in that we train our best models to the end, but optimize efficiency by discarding unpromising models early on. However, it is critically different from comparable algorithms such as Hyperband and Successive Halving \cite{jamieson16} in that it allows the evolution algorithm to dynamically select new promising candidates as the search progresses; Hyperband and Successive Halving establish their candidate pool a priori, which we demonstrate is ineffective in our large search space in Section 5.

\section{Methods}
We employ evolution-based architecture search because it is simple and has been shown to be more efficient than reinforcement learning when resources are limited \cite{real19}. We use the same \textit{tournament selection}~\cite{goldberg91} algorithm as Real et al. \yrcite{real19}, with the aging regularization omitted, and so encourage the reader to view their in-depth description of the method. In the interest of saving space, we will only give a brief overview of the algorithm here. 

Tournament selection evolutionary architecture search is conducted by first defining a \textit{gene encoding} that describes a neural network architecture; we describe our encoding in the following Search Space subsection. An initial \textit{population} is then created by randomly sampling from the space of gene encodings to create \textit{individuals}. These individuals, each corresponding to a neural architecture, are trained and assigned \textit{fitnesses}, which in our case are the models' negative log perplexities on the WMT'14 En-De \textit{validation set}. The population is then repeatedly \textit{sampled} from to produce \textit{subpopulations}, from which the individual with the highest fitness is selected as a \textit{parent}. Selected parents have their gene encodings \textit{mutated} -- encoding fields randomly changed to different values -- to produce \textit{child models}. These child models are then assigned a fitness via training and evaluating on the target task, as the initial population was. When this fitness evaluation concludes, the population is sampled from once again, and the individual in the subpopulation with the lowest fitness is \textit{killed}, meaning it is removed from the population. The newly evaluated child model is then added to the population, taking the killed individual's place. This process is repeated and results in a population with high fitness individuals, which in our case represent well-performing architectures.

\subsection{Search Space}
Our encoding search space is inspired by the NASNet search space \cite{zoph18}, but is altered to allow it to express architecture characteristics found in recent state-of-the-art feed-forward seq2seq networks.
Crucially, we ensured that the search space can represent the Transformer, so that we could seed the initial population with it.

\begin{figure}[h!]
\vspace{-0.1in}
\begin{center}
\centerline{\includegraphics[width=0.8\columnwidth]{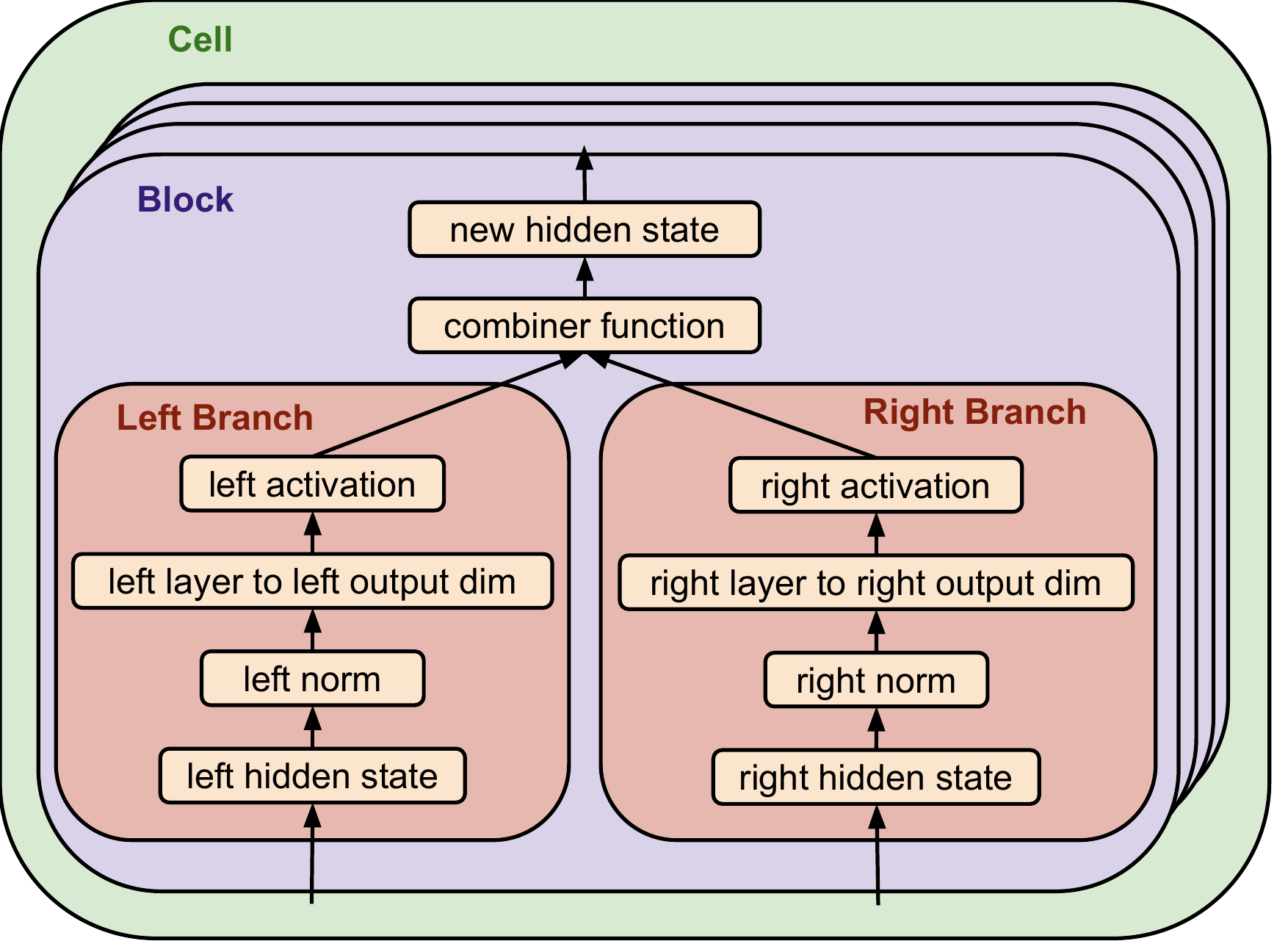}}
\caption{\textbf{Architecture composition from encoding.} Each block produces a new hidden state that is added to the pool of hidden states subsequent blocks can select as branch inputs. Each encoder has 6 unique blocks per cell and each decoder has 8 unique blocks per cell. Each cell is repeated \textit{number of cells} times.}
\label{fig:ss}
\end{center}
\vspace{-0.25in}
\end{figure}

Our search space consists of two stackable cells, one for the model encoder and one for the decoder (see Figure~\ref{fig:ss}). Each cell contains NASNet-style blocks, which receive two hidden state inputs and produce new hidden states as outputs; the encoder contains six blocks and the decoder contains eight blocks, so that the Transformer can be represented exactly. The blocks perform separate transformations to each input and then combine the transformation outputs together to produce a single block output; we will refer to the transformations applied to each input as a \textit{branch}. Our search space contains five branch-level search fields (input, normalization, layer, output dimension and activation), one block-level search field (combiner function) and one cell-level search field (number of cells).

In our search space, a child model's genetic encoding is expressed as: $[$\textit{left input}, \textit{left normalization}, \textit{left layer}, \textit{left relative output dimension}, \textit{left activation}, \textit{right input}, \textit{right normalization}, \textit{right layer}, \textit{right relative output dimension}, \textit{right activation}, \textit{combiner function}$]$ $\times$ 14 + $[$\textit{number of cells}$]$ $\times$ 2, with the first 6 blocks allocated to the encoder and the latter 8 allocated to the decoder. Given the vocabularies described in the Supplementary Materials, this yields a search space of $7.30 * 10^{115}$ models, although we do shrink this to some degree by introducing constraints (see the Supplementary Materials for more details).

\subsection{Seeding the Search Space with Transformer}

 While previous neural architecture search works rely on well-formed hand crafted search spaces \cite{zoph18}, we intentionally leave our space minimally tuned, in a effort to alleviate our manual burden and emphasize the role of the automated search method. To help navigate the large search space, we find it easier to warm start the search process by seeding our initial population with a known strong model, in this case the Transformer. This anchors the search to a known good starting point and guarantees at least a single strong potential parent in the population as the generations progress. We offer empirical support for these claims in our Results section. 

\subsection{Evolution with Progressive Dynamic Hurdles}

The evolution algorithm we employ is adapted from the tournament selection evolutionary architecture search proposed by Real et al. \yrcite{real19}, described above. Unlike Real et al. \yrcite{real19} who conducted their search on CIFAR-10, our search is conducted on a task that takes much longer to train and evaluate on. 
Specifically, to train a Transformer to peak performance on WMT'14 En-De requires $\sim$300K training steps, or 10 hours, in the base size when using a single Google TPU V.2 chip, as we do in our search. In contrast, Real et al. \yrcite{real19} used the less resource-intensive CIFAR-10 task \cite{krizhevsky09}, which takes about two hours to train on, to assess their models during their search, as it was a good proxy for ImageNet \cite{deng2009} performance \cite{zoph18}. However, in our preliminary experimentation we could not find a proxy task that gave adequate signal for how well each child model would perform on the full WMT'14 En-De task; we investigated using only a fraction of the data set and various forms of aggressive early stopping.

To address this problem we formulated a method to dynamically allocate resources to more promising architectures according to their fitness. This method, which we refer to as \textit{Progressive Dynamic Hurdles} (PDH), allows models that are consistently performing well to train for more steps. It begins as ordinary tournament selection evolutionary architecture search with early stopping, with each child model training for a relatively small $s_0$ number of steps before being evaluated for fitness. However, after a predetermined number of child models, $m$, have been evaluated,  a \textit{hurdle}, $h_0$, is created by calculating the the mean fitness of the current population. For the next $m$ child models produced, models that achieve a fitness greater than $h_0$ after $s_0$ train steps are granted an additional $s_1$ steps of training and then are evaluated again to determine their final fitness. Once another $m$ models have been considered this way, another hurdle, $h_1$, is constructed by calculating the mean fitness of all members of the current population that were trained for the maximum number of steps. For the next $m$ child models, training and evaluation continues in the same fashion, except models with fitness greater than $h_1$ after $s_0 + s_1$ steps of training are granted an additional $s_2$ number of train steps, before being evaluated for their final fitness. This process is repeated until a satisfactory number of maximum training steps is reached. Algorithm 1 (Supplementary Materials) formalizes how the fitness of an individual model is calculated with hurdles and Algorithm 2 (Supplementary Materials) describes tournament selection augmented with Progressive Dynamic Hurdles.

\begin{figure}[h!]
\vspace{-0.2in}
\centerline{\includegraphics[width=0.6\columnwidth]{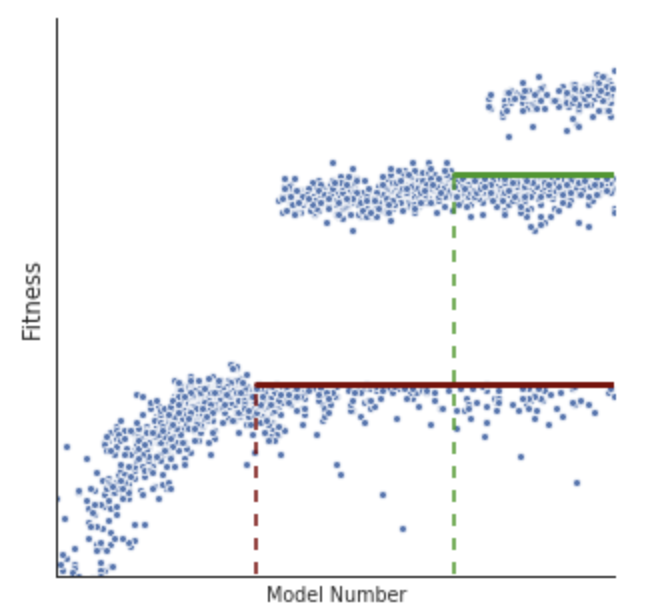}}
\caption{\textbf{Evolution architecture search with hurdles.} The y-axis represents architecture fitness and the x-axis represents the order in which candidate models were created. The solid purple and green lines represent the values of the first and second hurdles, respectively. The dashed purple and green lines represent the points at which each of the corresponding hurdles were introduced. Points to the left of the purple dashed line were generated using unaltered tournament selection. Between the purple and green dashed lines, models with a fitness above the solid purple line were granted additional train steps, forming a higher fitness cluster. To the right of the green dashed line, models with a fitness greater than the solid green line were granted a second round of additional train steps.} 
\vspace{-0.1in}
\end{figure}

Although different child models may train for different numbers of steps before being assigned their final fitness, this does not make their fitnesses incomparable. Tournament selection evolution is only concerned with relative fitness rank when selecting which subpopulation members will be killed and which will become parents; the margin by which one candidate is better or worse than the other members of the subpopulation does not matter. Assuming no model overfits during its training, which is what we observed in our experiments, and that its fitness monotonically increases with respect to the number of train steps it is allocated, a comparison between two child models can be viewed as a comparison between their fitnesses at the lower of the two's cumulative train steps. Since the model that was allocated more train steps performed, by definition, above the fitness hurdle for the lower number of steps and the model that was allocated less steps performed, by definition, at or below that hurdle at the lower number of steps, it is guaranteed that the model with more train steps was better when it was evaluated at the lower number of train steps.

The benefit of altering the fitness algorithm this way is that poor performing child models will not consume as many resources when their fitness is being computed. As soon as a candidate's fitness falls below a tolerable amount, its evaluation immediately ends. This may also result in good candidates being labeled as bad models if they are only strong towards the latter part of training. However, the resources saved as a result of discarding many bad models improves the overall quality of the search enough to justify potentially also discarding some good ones; this is supported empirically in our Results section.

\section{Experiment Setup}

\subsection{Datasets}

\paragraph{Machine Translation} We use three different machine translation datasets to perform our experiments, all of which were taken from their Tensor2Tensor implementations\footnote{\nohyph https://github.com/tensorflow/tensor2tensor/tree/master/ tensor2tensor/data\_generators}. The first is WMT English-German, for which we mimic Vaswani et al.'s \yrcite{vaswani17} setup, using WMT'18 En-De training data without ParaCrawl \cite{paracrawl18}, yielding 4.5 million sentence pairs. In the same fashion, we use newstest2013 for development and test on newstest2014. The second translation dataset is WMT En-Fr, for which we also replicate Vaswani et.al's \yrcite{vaswani17} setup. We train on the 36 million sentence pairs of WMT'14 En-Fr, validate on newstest2013 and test on newstest2014. The final translation dataset is WMT English-Czech (En-Cs). We used the WMT'18 training dataset, again without ParaCrawl, and used newstest2013 and newstest2014 as validation and test sets. For all tasks, tokens were split using a shared source-target vocabulary of about 32K word-pieces \cite{wu16}.

All datasets were generated using Tensor2Tensor's ``packed" scheme; sentences were shuffled and concatenated together with padding to form uniform 256 length inputs and targets, with examples longer than 256 being discarded. This yielded batch sizes of 4096 tokens per GPU or TPU chip; accordingly, 16 TPU chip configurations had $\sim$66K tokens per batch and 8 GPU chip configurations had $\sim$33K tokens per batch.

\paragraph{Language Modeling} For language modeling we used the 1 Billion Word Language Model Benchmark (LM1B) \cite{chelba13}, also using its ``packed" Tensor2Tensor implementation. Again the tokens are split into a vocabulary of approximately 32K word-pieces and the sentences are shuffled.


\subsection{Training Details and Hyperparameters}

\paragraph{Machine Translation} All of our experiments used Tensor2Tensor's Transformer TPU hyperparameter settings\footnote{\label{transformer_git}\nohyph https://github.com/tensorflow/tensor2tensor/blob/master/ tensor2tensor/models/transformer.py}. These are nearly identical to those used by Vaswani et al. \yrcite{vaswani17}, but modified to use the memory-efficient Adafactor \cite{shazeer18} optimizer. Aside from using the optimizer itself, these hyperparameters also set the warmup to a constant learning rate of $10^{-2}$ over 10K steps and then uses inverse-square-root learning-rate decay. For our experiments, we make only one change, which is to alter this decay so that it reaches 0 at the final step of training, which for our non-search experiments is uniformly 300K. We found that the our search candidate models, the Transformer, and the Evolved Transformer all benefited from this and so experimented with using linear decay, single-cycle cosine decay \cite{loshchilov17} and a modified inverse-square-root decay to 0 at 300K steps:  $lr = step^{-0.00303926} - .962392$; every decay was paired with the same constant $10^{-2}$ warmup. We used WMT En-De validation perplexity to gauge model performance and found that the Transformer preferred the modified inverse-square-root decay. Therefore, this is what we used for both all our Transformer trainings and the architecture searches themselves. The Evolved Transformer performed best with cosine decay and so that is what we used for all of its trainings. Besides this one difference, the hyperparameter settings across models being compared are exactly the same. Because decaying to 0 resulted in only marginal weight changes towards the end of training, we did not use checkpoint averaging, except where noted.

Per-task there is one additional hyperparameter difference, which is dropout rate. For ET and all search child models, dropout was applied uniformly after each layer, approximating the Transformer's more nuanced dropout scheme. For En-De and En-Cs, all ``big" and ``deep" sized models were given a higher dropout rate of 0.3, keeping in line with Vaswani et al. \yrcite{vaswani17}, and all other models with an input embedding size of 768 are given a dropout rate of 0.2. Aside from this, hyperparameters are identical across all translation tasks.

For decoding we used the same beam decoding configuration used by Vaswani et al. \yrcite{vaswani17}. That is a \textit{beam size} of 4, \textit{length penalty} ($\alpha$) of 0.6, and \textit{maximum output length} of input length + 50. All BLEU is calculated using case-sensitive tokenization\footnote{\nohyph https://github.com/moses-smt/mosesdecoder/blob/master/ scripts/generic/multi-bleu.perl} and for WMT'14 En-De we also use the compound splitting that was used in Vaswani et al. \yrcite{vaswani17}.

\paragraph{Language Modeling}
Our language model training setup is identical to our machine translation setup except we remove label smoothing and lower the intra-attention dropout rate to 0. This was taken from the Tensor2Tensor hyperparameters for LM1B\footnotemark[2].

\subsection{Search Configurations}
All of the architecture searches we describe were run on WMT'14 En-De. They utilized the search space and tournament selection evolution algorithm described in our Methods section. Unless otherwise noted, each search used 200 workers, which were equipped with a single Google TPU V.2 chip for training and evaluation. We maintained a population of size 100 with subpopulation sizes for both killing and reproducing set to 30. Mutations were applied independently per encoding field at a rate of 2.5\%. For fitness we used the negative log perplexity of the validation set instead of BLEU because, as demonstrated in our Results section, perplexity is more consistent and that reduced the noise of our fitness signal.

\section{Results}
In this section, we will first benchmark the performance of our search method, Progressive Dynamic Hurdles, against other evolutionary search methods~\cite{real17,real19}. We will then benchmark the Evolved Transformer, the result of our search method, against the Transformer~\cite{vaswani17}.  

\subsection{Ablation Study of Search Techniques}

We tested our evolution algorithm enhancements -- using PDH and warm starting by seeding the initial population with the Transformer -- against control searches that did not use these techniques; without our enhancements, these controls function the same way as Real et. al's \yrcite{real19} searches, without aging regularization. Each search we describe was run 3 times and the top model from each run was retrained on a single TPU V.2 chip for 300K steps. The performance of the models after retraining is given in Table \ref{table:hurdles_experiments}.

\vskip -0.1in
\begin{table}[!h]
\begin{center}
\begin{small}
\begin{sc}
\begin{tabular}{lcccc}
\toprule
Seed Model & \thead{Train \\ Steps} & \thead{Num \\ Models} & \thead{Top Model \\ Perplexity} \\
\midrule
Transformer    & PDH & 6000 & \textbf{4.50} $\pm$ 0.01 \\
Random         & PDH & 6000 & 5.23 $\pm$ 0.19 \\
Transformer    & 15K& 29714 & 4.57 $\pm$ 0.01 \\
Transformer    & 30K& 14857 & 4.53 $\pm$ 0.07 \\
Transformer    & 180K& 2477 & 4.58 $\pm$ 0.05 \\
Transformer    & 300K& 1486 & 4.61 $\pm$ 0.02 \\
\bottomrule
\end{tabular}
\end{sc}
\end{small}
\end{center}
\vskip -0.1in
\caption{\textbf{Top model validation perplexity of various search setups.} Number of models were chosen to equalize resource consumption.}
\vskip -0.1in
\label{table:hurdles_experiments}
\end{table}

\begin{figure}[th!]
\centerline{\includegraphics[width=\columnwidth,height=403pt]{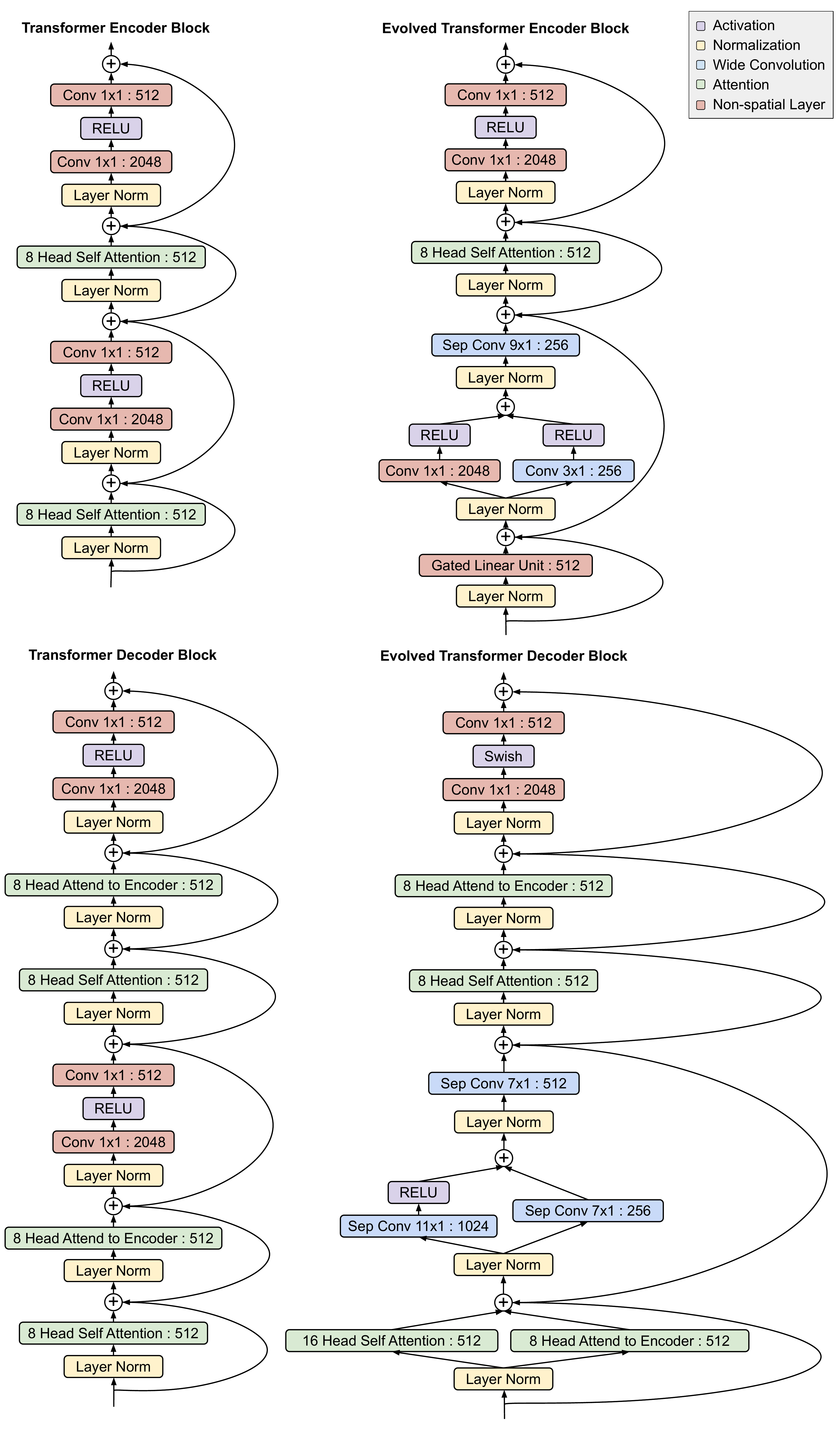}}
\vspace{-0.1in}
\caption{\textbf{Transformer and Evolved Transformer architecture cells.} The four most notable aspects of the found architecture are the use of 1) wide depth-wise separable convolutions, 2) Gated Linear Units~\cite{dauphin2017language}, 3) branching structures and 4) swish activations~\cite{ramachandran17}. Both the ET encoder and decoder independently developed a branched lower portion of wide convolutions. Also in both cases, the latter portion is almost identical to the Transformer.
}
\label{fig:architectures}
\vspace{-0.1in}
\end{figure}

Our proposed search (Table \ref{table:hurdles_experiments} row 1), which used both PDH and Transformer seeding, was run first, with hurdles created every 1K models ($m = 1000$) and six 30K train step (1 hour) increments ($s=<30, 30, 30, 30, 30, 30>$). To test the effectiveness of seeding with the Transformer, we ran an identical search that was instead seeded with random valid encodings (Table \ref{table:hurdles_experiments} row 2). To test the effectiveness of PDH, we ran four controls (Table 1 rows 3-6) that each used a fixed number of train steps for each child model instead of hurdles (Table 1 column 2). For these we used the step increments (30K), the maximum number of steps our proposed search ultimately reaches (180K), the total number of steps each top model receives when fully trained to gauge its final performance (300K), and half the step increments (15K), recognizing the gains from evaluating a larger number of models in the 30K steps control case. To determine the number of child models each of these searches would be able to train, we selected the value that would make the total amount of resources used by each control search equal to the maximum amount of resources used for our proposed searches, which require various amounts of resources depending on how many models fail to overcome hurdles. In the three trials we ran, our proposed search's total number of train steps used was 422M $\pm$ 21M, with a maximum of 446M. Thus the number of child models allotted for each non-PDH control search was set so that the total number of child model train steps used would be 446M.

As demonstrated in Table 1, the search we propose, with PDH and Transformer seeding, has the best performance on average. It also is the most consistent, having the lowest standard deviation. Of all the searches conducted, only a single control run -- ``30K no hurdles" (Table 1 row 3) -- produced a model that was better than any of our proposed search's best models. At the same time, the ``30K no hurdles" setup also produced models that were significantly worse, which explains its high standard deviation. This phenomenon was a chief motivator for our developing this method. Although aggressive early stopping has the potential to produce strong models for cheap, searches that utilize it can also venture into modalities in which top fitness child models are only strong early on; without running models for longer, whether or not this is happening cannot be detected. For example, the 15K search performed worse than the 30K setting, despite evaluating twice as many models. Although the 180K and 300K searches did have insight into long term performance, it was in a resource-inefficient manner that hurt these searches by limiting the number of generations they produced; for the ``180K no hurdles" run to train as many models as PDH would require 1.08B train steps, over double what PDH used in our worst case.

Searching with random seeding also proved to be ineffective, performing considerably worse than every other configuration. Of the five searches run, random seeding was the only one that had a top model perplexity higher than the Transformer, which is 4.75 $\pm$ 0.01 in the same setup.  

\vspace{-5pt}
\subsection{Main Search.} After confirming the effectiveness of our search procedure, we launched a larger scale version of our search using 270 workers. We trained 5K models per hurdle ($m=5000$) and used larger step increments to get a closer approximation to 300K step performance: $s = <60, 60, 120>$. The setup was the same as the Search Techniques experiments, except after 11K models we lowered the mutation rate to 0.01 and introduced the \textsc{NONE} value to the normalization mutation vocabulary.

The search ran for 15K child models, requiring a total of 979M train steps. Over 13K models did not make it past the first hurdle, drastically reducing the resources required to view the 240 thousandth train step for top models, which would have cost 3.6B train steps for the same number of models without hurdles. After the search concluded, we then selected the top 20 models and trained them for the full 300K steps, each on a single TPU V.2 chip. The model that ended with the best perplexity is what we refer to as the Evolved Transformer (ET). Figure \ref{fig:architectures} shows the ET architecture. The most notable aspect of the Evolved Transformer is the use of wide depth-wise separable convolutions in the lower layers of the encoder and decoder blocks. The use of depth-wise convolution and self-attention was previously explored in QANet~\cite{wei2018fast}, however the overall architectures of the Evolved Transformer and QANet are different in many ways: e.g., QANet has smaller kernel sizes and no branching structures. The performance and analysis of the Evolved Transformer will be shown in the next section.

\subsection{The Evolved Transformer: Performance and Analysis}

To test the effectiveness of the found architecture -- the Evolved Transformer -- we compared it to the Transformer in its Tensor2Tensor training regime on WMT'14 En-De. Table \ref{table:ende_gpu} shows the results of these experiments run on the same 8 NVIDIA P100 hardware setup that was used by Vaswani et al. \yrcite{vaswani17}.  Observing ET's improved performance at parameter-comparable ``base" and ``big" sizes, we were also interested in understanding how small ET could be shrunk while still achieving the same performance as the Transformer. To create a spectrum of model sizes for each architecture, we selected different input embedding sizes and shrank or grew the rest of the model embedding sizes with the same proportions. Aside from embedding depths, these models are identical at all sizes, except the ``big" 1024 input embedding size, for which all 8 head attention layers are upgraded to 16 head attention layers, as was done in Vaswani et al. \yrcite{vaswani17}.

\begin{table*}[h!]
\begin{center}
\begin{small}
\begin{sc}
\begin{tabular}{cccc@{\hskip 0.4in}cc@{\hskip 0.4in}cc}
\toprule
Task & Size & \thead{Tran \\ Params} & \thead{ET \\ Params} & Tran Perp  & ET Perp & Tran BLEU & ET BLEU \\
\midrule
WMT'14 En-De  & Base & 61.1M & 64.1M & 4.24 $\pm$ 0.03 & \textbf{4.03} $\pm$ 0.02 & 28.2 $\pm$ 0.2 & \textbf{28.4} $\pm$ 0.2 \\
WMT'14 En-De    & Big & 210.4M & 221.7M & 3.87 $\pm$ 0.02 & \textbf{3.77} $\pm$ 0.02 & 29.1 $\pm$ 0.1 & \textbf{29.3} $\pm$ 0.1 \\
WMT'14 En-De    & Deep & 224.0M & 218.1M & 3.86 $\pm$ 0.02 & \textbf{3.69} $\pm$ 0.01 & 29.2 $\pm$ 0.1 & \textbf{29.5} $\pm$ 0.1 \\
\midrule
WMT'14 En-Fr    & Base & 60.8 & 63.8M & 3.61 $\pm$ 0.01 & \textbf{3.42} $\pm$ 0.01 & 40.0 $\pm$ 0.1 & \textbf{40.6} $\pm$ 0.1 \\
WMT'14 En-Fr    & Big & 209.8M & 221.2M & 3.26 $\pm$ 0.01 & \textbf{3.13} $\pm$ 0.01 & 41.2 $\pm$ 0.1 & \textbf{41.3} $\pm$ 0.1 \\
\midrule
WMT'14 En-Cs    & Base & 59.8M & 62.7M & 4.98 $\pm$ 0.04 & \textbf{4.42} $\pm$ 0.01 & 27.0 $\pm$ 0.1 & \textbf{27.6} $\pm$ 0.2 \\
WMT'14 En-Cs    & Big & 207.6M & 218.9M & 4.43 $\pm$ 0.01 & \textbf{4.38} $\pm$ 0.03 & 28.1 $\pm$ 0.1 & \textbf{28.2} $\pm$ 0.1 \\ 
\midrule
LM1B            & Big & 141.1M & 151.8M & 30.44 $\pm$ 0.04 & \textbf{28.60} $\pm$ 0.03 & - & - \\ 
\bottomrule
\end{tabular}
\end{sc}
\end{small}
\end{center}
\caption{\textbf{Comparison between the Transformer and ET trained on 16 TPU V.2 chips.} For Translation, perplexity was calculated on the validation set and BLEU was calculated on the test set. For LM1B, perplexity was calculated on the test set. ET shows consistent improvement by at least one standard deviation on all tasks. It excels at the base size the search was conducted in, with an improvement of 0.6 BLEU in both En-Fr and En-Cs.}
\vskip -0.1in
\label{table:all_tasks}
\end{table*}

\begin{figure}[h!]
\begin{center}
\centerline{\includegraphics[width=0.6\columnwidth]{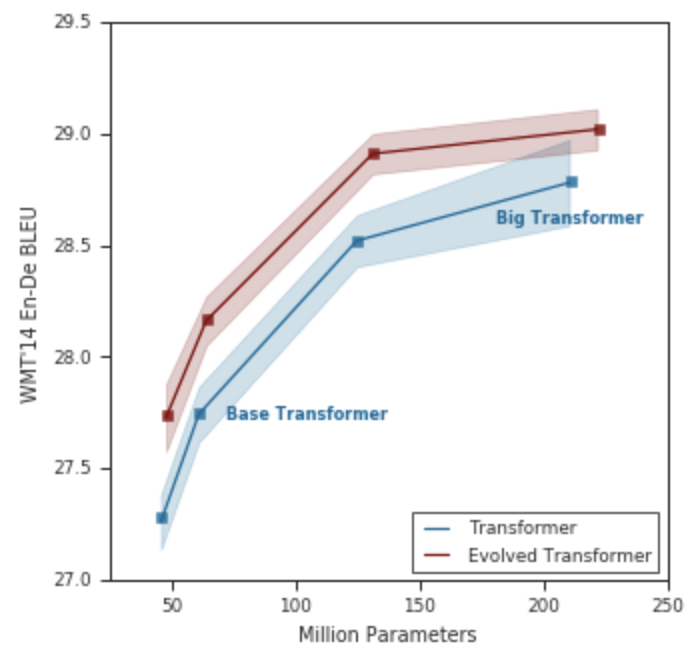}}
\caption{\textbf{Performance comparison of the Evolved Transformer against the Transformer across number of parameters.}
}
\vspace{-0.2in}
\label{fig:et}
\end{center}
\vspace{-0.2in}
\end{figure}

ET demonstrates stronger performance than the Transformer at all sizes, with the largest difference of 0.7 BLEU at the smallest, mobile-friendly, size of $\sim$7M parameters. Performance on par with the ``base" Transformer was reached when ET used just 78.4\% of its parameters and performance of the ``big" Transformer was exceeded by the ET model that used  37.6\% less parameters. Figure~\ref{fig:et} shows the FLOPS vs. BLEU performance of both architectures. 

\begin{table*}[h!]
\begin{center}
\begin{small}
\begin{tabular}{ccccccc}
\toprule
Model & \thead{Embedding \\ Size} & Parameters & Perplexity & BLEU & \thead{ $\Delta$ BLEU} \\
\midrule
Transformer & 128 & 7.0M & 8.62 $\pm$ 0.03 & 21.3 $\pm$ 0.1 & - \\
ET & 128 & 7.2M & \textbf{7.62} $\pm$ 0.02 & \textbf{22.0} $\pm$ 0.1 & \textbf{+ 0.7} \\
\midrule
Transformer & 432 & 45.8M & 4.65 $\pm$ 0.01 & 27.3 $\pm$ 0.1 & - \\
ET & 432 & 47.9M & \textbf{4.36} $\pm$ 0.01 & \textbf{27.7} $\pm$ 0.1 & + 0.4 \\
\midrule
Transformer & 512 & 61.1M & 4.46 $\pm$ 0.01 & 27.7 $\pm$ 0.1 & - \\
ET & 512 & 64.1M & \textbf{4.22} $\pm$ 0.01 & \textbf{28.2} $\pm$ 0.1 & + 0.5 \\
\midrule
Transformer & 768 & 124.8M & 4.18 $\pm$ 0.01 & 28.5 $\pm$ 0.1 & -\\
ET & 768 & 131.2M & \textbf{4.00} $\pm$ 0.01 & \textbf{28.9} $\pm$ 0.1 & + 0.4 \\
\midrule
Transformer & 1024 & 210.4M & 4.05 $\pm$ 0.01 & 28.8 $\pm$ 0.2 & - \\
ET & 1024 & 221.7M & \textbf{3.94} $\pm$ 0.01 & \textbf{29.0} $\pm$ 0.1 & + 0.2 \\
\bottomrule
\end{tabular}
\end{small}
\end{center}
\caption{\textbf{WMT'14 En-De comparison on 8 NVIDIA P100 GPUs.} Each model was trained 10 to 15 times, depending on resource availability. Perplexity is calculated on the validation set and BLEU is calculated on the test set.}
\label{table:ende_gpu}
\vspace{-0.25in}
\end{table*}

To test if ET's strong performance generalizes, we also compared it to the Transformer on an additional three well-established language tasks: WMT'14 En-Fr, WMT'14 En-Cs, and LM1B.\footnote{For LM1B, we only use the decoder architecture, with \textit{attend to encoder} layers removed.} Upgrading to 16 TPU V.2 chips, we doubled the number of synchronous workers for these experiments, pushing both models to their higher potential \cite{ott18}. We ran each configuration 3 times, except WMT En-De, which we ran 6 times; this was a matter of resource availability and we gave priority to the task we searched on. As shown in Table \ref{table:all_tasks}, ET performs at least one standard deviation above the Transformer in each of these tasks. Note that the Transformer mean BLEU scores in all of our experiments for WMT'14 En-Fr and WMT'14 En-De are higher than those originally reported by Vaswani et al. \yrcite{vaswani17}.

As can be seen in Tables \ref{table:ende_gpu} and \ref{table:all_tasks}, the Evolved Transformer is much more effective than the Transformer at smaller model sizes. 
At the ``big" model size, its BLEU performance saturates and the gap between the Evolved Transformer and the Transformer becomes smaller. One explanation for this behavior is that overfitting starts to occur at big model sizes, but we expect that data augmentation~\cite{ott18} or hyperparameter tuning could improve performance.
For example, we found that simply increasing the embedding size was not the best way to grow ET from the ``base" size we searched over to a larger size. Depth should also be tuned in conjunction with embedding size, when controlling for number of parameters. For both the Transformer and ET, we tried four additional embedding sizes, [512, 640, 768, 896], adjusting the depth accordingly so that all resulting models had a similar number of parameters. Using validation BLEU to determine the best configuration, we found that ET performed best with an embedding depth of 640, increasing its number of encoder cells from 3 to 9 and its number of decoder cells from 4 to 10. The Transformer also benefited from additional depth, although not to the same degree, achieving maximum performance at the 768 embedding size, with 6 encoder cells and 6 decoder cells. These results are included in Table \ref{table:all_tasks}, labeled as ``Deep" size.

\begin{table}
\begin{small}
\begin{tabular}{cccc}
\toprule
Model & Params & BLEU & \thead{SacreBLEU \\ \cite{post18}} \\
\midrule
Gehring et al. \yrcite{gehring2017} & 216M & 25.2 & - \\
Vaswani et al. \yrcite{vaswani17} & 213M & 28.4 & - \\
Ahmed et al. \yrcite{ahmed17} & 213M & 28.9 & - \\
Chen et al. \yrcite{chen18} & 379M & 28.5 & - \\
Shaw et al. \yrcite{shaw18} & 213M & 29.2 & - \\
Ott et al. \yrcite{ott18} & 210M & 29.3 & 28.6 \\
Wu et al. \yrcite{wu2018pay} & 213M & 29.7 & - \\
\midrule
Evolved Transformer & 218M & \textbf{29.8} & \textbf{29.2} \\
\bottomrule
\end{tabular}
\end{small}
\caption{\textbf{Model comparison on WMT'14 En-De.}}
\vskip -0.2in
\label{table:state_of_the_art}
\end{table}

\vskip -0.2in
To compare with other previous results, we trained the ET Deep model three times in our TPU setup on WMT'14 En-De, selected the best run according to validation BLEU and did a single decoding on the test set. We also copied previous state-of-the-art result setups by averaging the last 20 model checkpoints from training and decoding with a beam width of 5 \cite{vaswani17, ott18, wu2018pay}. As a result, the Evolved Transformer achieved a new state-of-the-art BLEU score of 29.8 (Table \ref{table:state_of_the_art}).

\section{Conclusion}

We presented the first neural architecture search conducted to find improved feed-forward sequence models. We first constructed a large search space inspired by recent advances in seq2seq models and used it to search directly on the computationally intensive WMT En-De translation task. To mitigate the size of our space and the cost of training child models, we proposed using both our Progressive Dynamic Hurdles method and warm starting, seeding our initial population with a known strong model, the Transformer.

When run at scale, our search found the Evolved Transformer. In a side by side comparison against the Transformer in an identical training regime, the Evolved Transformer showed consistent stronger performance on both translation and language modeling. On the task we searched over, WMT'14 En-De, the Evolved Transformer established a new state-of-the-art of 29.8 BLEU. It also proved to be efficient at smaller sizes, achieving the same quality as the original "big" Transformer with 37.6\% less parameters and outperforming the Transformer by 0.7 BLEU at a mobile-friendly model size of $\sim$7M parameters.

\clearpage

\section*{Acknowledgements}

We would like to thank Ashish Vaswani, Jakob Uszkoreit, Niki Parmar, Noam Shazeer, Lukasz Kaiser and Ryan Sepassi for their help with Tensor2Tensor and for sharing their understanding of the Transformer. We are also grateful to David Dohan, Esteban Real, Yanping Huang, Alok Aggarwal, Vijay Vasudevan, and Chris Ying for lending their expertise in architecture search and evolution.

\bibliography{main}
\bibliographystyle{icml2019}

\newpage
\appendix

\section{Search Algorithms}
In the following, we describe the algorithm that we use to calculate child model fitness with hurdles (Algorithm~\ref{algo:1}) and evolution architecture search with Progressive Dynamic Hurdles (Algorithm~\ref{algo:2}).
\begin{algorithm} [h!]
   \caption{Calculate Model Fitness with Hurdles}
   \label{algo:1}
\begin{algorithmic}
  \STATE {\bfseries inputs:}
  \STATE\hspace{3pt} $model$: the child model
  \STATE\hspace{3pt} $s$: vector of train step increments
  \STATE\hspace{3pt} $h$: queue of hurdles
  \STATE
  \STATE append $\infty$ to $h$
  \STATE TRAIN\_N\_STEPS($model$, $s_0$)
  \STATE $fitness \gets$ EVALUATE($model$)
  \STATE $i \gets 0$
  \STATE $hurdle \gets h_i$
  \STATE
  \WHILE{$fitness > hurdle$}
  \STATE $i \gets i + 1$
  \STATE TRAIN\_N\_STEPS($model$, $s_i$)
  \STATE $fitness \gets$ EVALUATE($model$)
  \STATE $hurdle \gets h_{i}$
  \ENDWHILE
  \STATE {\bfseries return} $fitness$
\end{algorithmic}
\end{algorithm}

\paragraph{Algorithm 1} Calculating fitness with hurdles takes as arguments a child model, a vector of train step increments ($s$) and a queue of hurdles($h$). The child model is the candidate model in our neural architecture search. The vector of step increments describes the number of steps between each hurdle; its length must be greater than 0. The queue of hurdles describes what hurdles have already been established; its length must be in [0, $length(s)$).

The algorithm starts by first training the child model a fixed number of $s_0$ steps and evaluating on the validation set to produce a fitness, as is done in Real et al. \yrcite{real19}. After this baseline fitness is established, the hurdles ($h$) are compared against to determine if training should continue. Each $h_i$ denotes the fitness a child model must have after $\sum_{j=0}^i s_j$ train steps to continue training. Each time a hurdle $h_i$ is passed, the model is trained an additional $s_{i+1}$ steps. If the model's fitness ever falls below the hurdle corresponding to the number of steps it was trained for, training ends immediately and the current fitness is returned. If the model never falls below a hurdle and all hurdles have been passed, the child model receives one final training of $s_{length(h)}$ steps before fitness is returned; this is expressed in Algorithm 1 with $\infty$ being appended to the end of $h$.
\vskip 0.2in

\begin{algorithm} [h!]
   \caption{Evolution Architecture Search with PDH}
   \label{algo:2}
\begin{algorithmic}
  \STATE {\bfseries inputs:}
  \STATE\hspace{3pt} $s$: vector of train step increments
  \STATE\hspace{3pt} $m$: number of child models per hurdle
  \STATE
  \STATE $h \gets $ \textit{empty queue} 
  \STATE $i \gets 0$
  \STATE $population \gets$ INITIAL\_POPULATION()
  \STATE
  \WHILE{$i < $ LENGTH($s$) - 1}
  \STATE $population \gets$ EVOL\_N\_MODELS($population$,
  \STATE\hspace{150pt}$m$, $s$, $h$)
  \STATE $hurdle \gets$ MEAN\_FITNESS\_OF\_MAX($population$)
  \STATE append $hurdle$ to $h$
  \ENDWHILE
  \STATE
  \STATE $population \gets$ EVOL\_N\_MODELS($population$,
  \STATE\hspace{150pt}$m$, $s$, $h$)
  \STATE {\bfseries return} $population$
\end{algorithmic}
\end{algorithm}

\paragraph{Algorithm 2} Evolution architecture search with PDH takes as arguments a vector of train step increments ($s$) and a number of child models per hurdle ($m$). It begins as Real et al.'s \yrcite{real19} evolution architecture search with a fixed number of child model train steps, $s_0$. However, after $m$ child models have been produced, a hurdle is created by taking the mean fitness of the current population and it is added to the hurdle queue, $h$. Algorithm 1 is used to compute each child model's fitness and so if they overcome the new hurdle they will receive more train steps. This process is continued, with new hurdles being created using the mean fitness of all models that have trained the maximum number of steps and $h$ growing accordingly. The process terminates when $length(s) -1$ hurdles have been created and evolution is run for one last round of $m$ models, using all created hurdles.

\section{Search Space Information}
In our search space, a child model's genetic encoding is expressed as: $[$\textit{left input}, \textit{left normalization}, \textit{left layer}, \textit{left relative output dimension}, \textit{left activation}, \textit{right input}, \textit{right normalization}, \textit{right layer}, \textit{right relative output dimension}, \textit{right activation}, \textit{combiner function}$]$ $\times$ 14 + $[$\textit{number of cells}$]$ $\times$ 2, with the first 6 blocks allocated to the encoder and the latter 8 allocated to the decoder. In the following, we will describe each of the components.

\paragraph{Input.} The first branch-level search field is \textit{input}. This specifies what hidden state in the cell will be fed as input to the branch. For each $i^{th}$ block, the input vocabulary of its branches is $[0, i)$, where the $j^{th}$ hidden state corresponds to the $j^{th}$ block output and the $0^{th}$ hidden state is the cell input.   

\paragraph{Normalization.} The second branch-level search field is \textit{normalization}, which is applied to each input before the layer transformation is applied. The normalization vocabulary is [\textsc{layer normalization} \cite{ba16}, \textsc{none}].

\paragraph{Layers.} The third branch-level search field is \textit{layer}, which is the neural network layer applied after the normalization. It's vocabulary is:
\begin{itemize}[noitemsep,nolistsep]
    \item \textsc{standard conv }$w$x1: for $w \in \{1, 3\}$
    \item \textsc{depthwise separable conv }$w$x1: for $w \in \{3, 5, 7, 9, 11\}$
    \item \textsc{lightweight conv }$w$x1 $r$: for $w \in \{3, 5, 7, 15\}$, $r \in \{1, 4, 16\}$ \cite{wu2018pay}. $r$ is the reduction factor, equivalent to $d/H$ described in Wu et al. \yrcite{wu2018pay}.
    \item $h$\textsc{ head attention}: for $h \in \{4, 8, 16\}$
    \item \textsc{gated linear unit}\cite{dauphin2017language}
    \item \textsc{attend to encoder}: (Only available to decoder)
    \item \textsc{identity}: No transformation applied to input
    \item \textsc{dead branch}: No output
\end{itemize}

For decoder convolution layers the inputs are shifted by $(w - 1) / 2$ so that positions cannot ``see" later predictions.

\paragraph{Relative Output Dimension.}
The fourth branch-level search field is \textit{relative output dimension}, which describes the output dimension of the corresponding layer. The Transformer is composed mostly of layers that project to the original input embedding depth (512 in the ``base" configuration), but also contains 1x1 convolutions that project up to a dimension of 4 times that depth (2048 in the ``base" configuration). The relative output dimension search field accounts for this variable output depth. It's vocabulary consists of 10 relative output size options: $[1, 10]$.

Here ``relative" refers to the fact that for every layer $i$ and $j$, each of their absolute output dimensions, $a$, and relative output dimensions, $d$, will obey the ratio: $a_i/a_j  = d_i/d_j$. We determine the absolute output dimensions for each model by finding a scaling factor, $s$, such that for every layer $i$, $a_i = d_i * s$ and the resulting model has an appropriate number of parameters; at the end of this section, we  describe our constraints on number of model parameters. There may be multiple values of $s$ for any one model that satisfy this constraint, and so for our experiments we simply perform a binary search and use the first valid value found. If no valid value is found, we reject the child model encoding as invalid and produce a new one in its stead.

We chose a vocabulary of relative sizes instead of absolute sizes because we only allow models within a fixed parameter range, as described later in this section (Constraints). Using relative sizes allows us to increase the number of configurations that represent valid models in our search space, because we can dynamically shrink or grow a model to make it fit within the parameter bounds. We found that using absolute values, such as $[256, 512, 1024, 2048]$, increases the number of rejected models and thereby decreases the possible models that can be expressed.

This relative output dimensions field is ignored for both the \textsc{identity} and \textsc{dead branch} layers.

\paragraph{Activations.} The final branch-level search field is \textit{activation}, which is the non-linearity applied on each branch after the neural network layer. The activation vocabulary is \{\textsc{swish} \cite{ramachandran17,elfwing2018sigmoid}, \textsc{relu}, \textsc{leaky\_relu~\cite{maas2013rectifier}}, \textsc{none}\}.

\paragraph{Combiner Functions.} The block-level search field, \textit{combiner function}, describes how the left and right layer branches are combined together. Its vocabulary is \{\textsc{addition}, \textsc{concatenation}, \textsc{multiplication}\}. For \textsc{multiplication} and \textsc{addition}, if the right and left branch outputs have differing embedding depths, then the smaller of the two is padded so that the dimensionality matches. For \textsc{addition} the padding is 0's; for \textsc{multiplication} the padding is 1's.

\paragraph{Number of Cells.} The cell-level search field is \textit{number of cells} and it describes the number of times the cell is repeated. Its vocabulary is [1,6].

\paragraph{Composition.}
Each child model is defined by two cells, one for the encoder and one for the decoder. The encoder cell contains 6 blocks and the decoder cell contains 8 blocks. Each block contains two branches, each of which takes a previous hidden layer as input, and then applies its normalization, layer (with specified relative output dimensions) and activation to it. The two branches are then joined with the combiner function. Any unused hidden states are automatically added to the final block output via addition. Both the encoder and decoder cells defined this way are repeated their corresponding \textit{number of cells} times and connected to the input and output embedding portions of the network to produce the final model; we use the same embedding scheme described by Vaswani et al. \yrcite{vaswani17} for all models. See Figure \ref{fig:ss} for a depiction of this composition.

\paragraph{Constraints.}
In the interest of having a fair comparison across child models, we limit our search to only architectures configurations that can contain between 59.1 million and 64.1 million parameters when their relative output dimensions are scaled; in the Tensor2Tensor \cite{vaswani18} implementation we use, the base Transformer has roughly 61.1 million parameters on WMT En-De, so our models are allowed 3 million less or more parameters than that. Models that cannot be represented within this parameter range are not included in our search space.

Additionally, in preliminary experiment runs testing the effectiveness of our search space, we discovered three trends that hurt performance in almost every case. Firstly and most obviously is when a proposed decoder contains no \textsc{attend to encoder} layers. This results in the decoder receiving no signal from the encoder and thus the model output will not be conditioned on the input. Therefore, any model that does not contain \textsc{attend to encoder} is not in our search space. The second trend that we noticed was that models that had layer normalization removed were largely worse than their parent models. For this reason, we remove \textsc{none} from the \textit{normalization} mutation vocabulary for each experiment, unless otherwise specified. Lastly, we observed that an important feature of good models was containing an unbroken residual path from inputs to outputs; in our search space, this means a path of \textsc{identity} layers from cell input to output that are combined with \textsc{addition} at every \textit{combination function} along the way. Our final constraint is therefore that models that do not have unbroken residual paths from cell inputs to outputs are not in our search space.

\section{Ablation Study of the Evolved Transformer}
To understand what mutations contributed to ET's improved performance we conducted two rounds of ablation testing. In the first round, we began with the Transformer and applied each mutation to it individually to measure the performance change each mutation introduces in isolation. In the second round, we began with ET and removed each mutation individually to again measure the impact of each single mutation. In both cases, each model was trained 3 times on WMT En-De for 300k steps with identical hyperparameters, using the inverse-square-root decay to 0 that the Transformer prefers. Each training was conducted on a single TPU V.2 chip. The results of these experiments are presented in Table~\ref{tab:ablation}; we use validation perplexity for comparison because it was our fitness metric.

In all cases, the augmented ET models outperformed the the augmented Transformer models, indicating that the gap in performance between ET and the Transformer cannot be attributed to any single mutation. The mutation with the seemingly strongest individual impact is the increase from 3 to 4 decoder cells. However, even when this mutation is introduced to the Transformer and removed from ET, the resulting augmented ET model still has a higher fitness than the augmented Transformer model.

To highlight the impact of each augmented model's mutation, we present not only their perplexities but also the difference between their mean perplexity and their unaugmented base model mean perplexity in the "Mean Diff" columns:

\textit{base model mean perplexity} - \textit{augmented mean perplexity}

This delta estimates the change in performance each mutation creates in isolation.
Red highlighted cells contain evidence that their corresponding mutation hurt overall performance. Green highlighted cells contain evidence that their corresponding mutation helped overall performance.

In half of the cases, both the augmented Transformer's and the augmented Evolved Transformer's performances indicate that the mutation was helpful. Changing the number of attention heads from 8 to 16 was doubly indicated to be neutral and changing from 8 head self attention to a GLU layer in the decoder was doubly indicated to have hurt performance. However, this and other mutations that seemingly hurt performance may have been necessary given how we formulate the problem: finding an improved model with a comparable number of parameters to the Transformer. For example, when the Transformer decoder cell is repeated 4 times, the resulting model has 69.6M parameters, which is outside of our allowed parameter range. Thus, mutations that shrank ET's total number of parameters, even at a slight degradation of performance, were necessary so that other more impactful parameter-expensive mutations, such as adding an additional decoder cell, could be used.

Other mutations have inconsistent evidence about how useful they are. This ablation study serves only to approximate what is useful, but how effective a mutation is also depends on the model it is being introduced to and how it interacts with other encoding field values. 

\definecolor{Green}{rgb}{0.7,0.9,0.55}
\definecolor{Red}{rgb}{1,0.65,0.65}
\renewcommand\theadfont{\tiny\scshape}
\begin{table*}[ht]
\begin{center}
\begin{tiny}
\begin{sc}
\begin{tabular}{ccccccccc}
\toprule
Mutation Field & \thead{Mutation \\ Block Index} & \thead{Mutation \\ Branch} & Transformer Value & ET Value & \thead{Transformer \\ Perplexity} & ET Perplexity & \thead{Transformer \\ Mean Diff} & \thead{ET \\ Mean Diff} \\
\midrule
decoder activation & 6 & left & RELU & Swish & 4.73 $\pm$ 0.01 & 4.51 $\pm$ 0.02 & \cellcolor{Green}-0.02 & \cellcolor{Green}0.04 \\
decoder activation & 2 & right & RELU & None & 4.73 $\pm$ 0.01 & 4.48 $\pm$ 0.00 & \cellcolor{Green} -0.02 & \cellcolor{Green}0.02 \\
decoder input & 1 & left & 1 & 0 & 4.74 $\pm$ 0.04 & 4.46 $\pm$ 0.00 & \cellcolor{Green}-0.01 & \cellcolor{Red}-0.01 \\
decoder layer & 0 & left & 8 head attention & 16 head attention & 4.75 $\pm$ 0.01 & 4.47 $\pm$ 0.01 & 0.0 & 0.0 \\
decoder layer & 2 & left & standard conv 1x1 & separable conv 11x1 & 4.67 $\pm$ 0.01 & 4.55 $\pm$ 0.00 &  \cellcolor{Green}-0.08 & \cellcolor{Green} 0.09 \\
decoder layer & 3 & left & standard conv 1x1 & separable conv 7x1  & 4.72 $\pm$ 0.01 & 4.46 $\pm$ 0.01 &  \cellcolor{Green} -0.03 & 0.0 \\
decoder layer & 2 & right & dead branch & separable conv 7x1 & 4.71 $\pm$ 0.02 & 4.47 $\pm$ 0.00 &  \cellcolor{Green}-0.04 &  \cellcolor{Green}0.01 \\
decoder norm & 3 & left & none & layer norm & 4.73 $\pm$ 0.00 & 4.45 $\pm$ 0.01 &  \cellcolor{Green}-0.02 &  \cellcolor{Red}-0.01 \\
decoder norm & 7 & left & none & layer norm & 4.73 $\pm$ 0.02 & 4.47 $\pm$ 0.02 &  \cellcolor{Green}-0.02 & \cellcolor{Green}0.01 \\
decoder output dim & 2 & left & 8 & 4 & 4.74 $\pm$ 0.01 & 4.45 $\pm$ 0.01 & \cellcolor{Green}-0.01 & \cellcolor{Red}-0.02 \\
decoder num cells & - & - & 3 & 4 & 4.62 $\pm$ 0.00 & 4.59 $\pm$ 0.01 & \cellcolor{Green}-0.13 & \cellcolor{Green} 0.12 \\
encoder layers & 0 & left & 8 head attention & gated linear unit & 4.80 $\pm$ 0.03 & 4.45 $\pm$ 0.02 & \cellcolor{Red} 0.05 & \cellcolor{Red} -0.01 \\
encoder layers & 2 & left & standard conv 1x1 & separable conv 9x1 & 4.69 $\pm$ 0.01 & 4.50 $\pm$ 0.00 & \cellcolor{Green} -0.06 & \cellcolor{Green} 0.04 \\
encoder layers & 1 & right & dead branch & standard conv 3x1 & 4.73 $\pm$ 0.01 & 4.47 $\pm$ 0.03 &  \cellcolor{Green}-0.02 & \cellcolor{Green}0.01 \\
encoder norms & 2 & left & none & layer norm & 4.79 $\pm$ 0.03 & 4.46 $\pm$ 0.02 & \cellcolor{Red} 0.04 & 0.0 \\
encoder output dim & 2 & left & 2 & 1 & 4.74 $\pm$ 0.01 & 4.45 $\pm$ 0.0 & \cellcolor{Green}-0.01 & \cellcolor{Red} -0.01 \\
\midrule
\bottomrule
\end{tabular}
\end{sc}
\end{tiny}
\end{center}
\caption{\textbf{Mutation Ablations}: Each mutation is described by the first 5 columns. The augmented Transformer and augmented ET perplexities on the WMT'14 En-De validation set are given in columns 6 and 7. Columns 7 and 8 show the difference between the unaugmented base model perplexity mean and the augmented model perplexity mean. Red highlighted cells indicate evidence that the corresponding mutation hurts overall performance. Green highlighted cells indicate evidence that the corresponding mutation helps overall performance.}
\label{tab:ablation}
\end{table*}

\end{document}